\documentclass{article}
\usepackage{spconf,amsmath,graphicx,lipsum, hyperref}
\usepackage[utf8]{inputenc} 
\usepackage[T1]{fontenc}    
\usepackage{hyperref}       
\usepackage{url}            
\usepackage{booktabs}       
\usepackage{amsfonts}       
\usepackage{nicefrac}       
\usepackage{microtype}      
\usepackage{amsmath,amsfonts,amssymb}       
\usepackage{bm}
\usepackage{graphicx}
\usepackage{subfig}
\usepackage{hhline}
\usepackage{xspace}
\usepackage[dvipsnames]{xcolor}
\usepackage{cite}
\graphicspath{{imgs/}}
\usepackage{epstopdf}
\usepackage{url}

\title{Graph Neural Network for Large-scale Network Localization}
\name{Wenzhong Yan$^{\star}$, Di Jin$^{\dagger}$, Zhidi Lin$^{\star}$ and Feng Yin$^{\star}$ \thanks{This work was supported by the National Natural Science Foundation of China under Grants 61701426. Corresponding author: Feng Yin.}}
  
\address{$^{\star}$  School of Science and Engineering, The Chinese University of Hong Kong, Shenzhen, China \\
$^{\dagger}$  Signal Processing Group, Technische Universitat Darmstadt, Darmstadt, Germany}

\begin{document}
\ninept 
\maketitle
\begin{abstract}
Graph neural networks (GNNs) are popular to use for classifying structured data in the context of machine learning. But surprisingly, they are rarely applied to regression problems. In this work, we adopt GNN for a classic but challenging nonlinear regression problem, namely the network localization. Our main findings are in order. First, GNN is potentially the best solution to large-scale network localization in terms of accuracy, robustness and computational time. Second, proper thresholding of the communication range is essential to its superior performance. Simulation results corroborate that the proposed GNN based method outperforms all state-of-the-art benchmarks by far. Such inspiring results are theoretically justified in terms of data aggregation, non-line-of-sight (NLOS) noise removal and low-pass filtering effect, all affected by the threshold for neighbor selection. Code is available at \url{https://github.com/Yanzongzi/GNN-For-localization}.
\end{abstract}
\begin{keywords}
Graph neural networks, large-scale, network localization, non-line-of-sight, thresholding.
\end{keywords}
\section{Introduction}
In recent years, graph neural networks~(GNNs) have achieved many state-of-the-art results in various graph-related learning tasks such as node classification, link prediction and graph classification \cite{kipf2016semi,hamilton2017inductive, velivckovic2017graph, xu2018powerful}. Comparing with the multi-layer perceptrons~(MLPs), GNNs can exploit extra information from the edges. More concretely, each node aggregates information of its adjacent nodes instead of merely using its own \cite{ortega2018graph, gama2018convolutional}. Though being effective, GNN models are often used to deal with classification tasks. While regression problems are more challenging and constitute a larger body of practical signal processing applications. In this paper, we consider a classic yet challenging nonlinear regression problem, namely network localization \cite{Patwari03}.

Network localization requires not only the measurements between agent nodes and anchor nodes, but also the measurements between the
agent nodes themselves. In the past decades, a variety of canonical network localization methods have been developed, including: 1) maximum likelihood estimation based methods \cite{Patwari03, Simonetto_Leus_2014}; 2) least-squares estimation based methods \cite{Wymeersch2009}; 3) multi-dimensional scaling based methods \cite{Costa2006}; 4) mathematical programming based methods \cite{Biswas_Ye_SDP, Tseng2007} and 5) Bayesian message passing based methods \cite{Ihler2005, Wymeersch2009, Jin20bayes}. 

Ignoring the effect due to non-line-of-sight (NLOS) propagation will incur severe performance degradation of the aforementioned methods. For remedy, one could perform NLOS identification for each link and then either discard the NLOS measurements or suppress them robustly in the aforementioned methods \cite{Win_2010_NLOS_UWB}. However, accurate NLOS identification requires large-scale offline calibration and huge amount of manpower. The NLOS effect can also be dealt with from algorithmic aspect. By assuming that the NLOS noise follows a certain probability distribution, the maximum likelihood estimation based methods were developed in \cite{Yin_ECM, Chen_Wang_So_Poor_2012}. However, model mismatch may cause severe performance degradation. In a recent work \cite{jin2020exploiting}, network localization problem is formulated as a regularized optimization problem in which the NLOS-inducing sparsity of the ranging-bias parameters was additionally exploited. Unfortunately, all of these methods are computationally expensive for large-scale networks.

In this paper, we propose a fresh GNN based network localization method that is able to achieve all desired merits at the same time. First, it provides extremely stable and highly accurate localization accuracy despite of severe NLOS propagations. Second, it does not require laborious offline calibration nor NLOS identification. Third, it is scalable to large-scale networks at an affordable computation cost. As far as we know, this is the first time that GNN has been applied to network localization.

The remainder of this paper is organized as follows. Our problem is first formulated in Section \ref{problem_formulation}. In Section \ref{GCN_reg}, we introduce a fresh GNN framework for network localization. Numerical results are provided in Section \ref{experiment}, followed by the theoretical performance justification in Section \ref{analysis}. Finally, we conclude the paper in Section \ref{conclusion}.

\section{Problem Formulation}\label{problem_formulation}
We consider a wireless network in two-dimensional (2-D) space, and extension to the 3-D case is straightforward. We let, without loss of generality, $\mathcal{S}_a = \{1,2,\dots,N_l\}$ be the set of indices of the anchors, whose positions $\mathbf{p}_i=[x_i,y_i]^\top, i\in \mathcal{S}_a$ are known, and $\mathcal{S}_b = \{N_l+1, N_l+2,\ldots,N\}$ be the set of indices of the agents, whose positions are unknown. A distance measurement made between any two nodes $i$ and $j$ is given by
\begin{equation}
	x_{ij} = d(\mathbf{p}_i, \mathbf{p}_j) + n_{ij},
\end{equation}
where $d(\mathbf{p}_i,\mathbf{p}_j):=\|\mathbf{p}_i-\mathbf{p}_j \|$ is the Euclidean distance and $n_{ij}$ is an additive measurement error due to line-of-sight (LOS) and NLOS propagation. 
A distance matrix, denoted by $\mathbf{X}\in\mathbb{R}^{N \times N}$, is constructed by stacking the distance measurements, where $x_{ij}$ is the $(ij)$-th entry of $\mathbf{X}$. Notably, the distance matrix $\mathbf{X}$ is a ``zero-diagonal'' matrix, because $x_{ii} = 0$ for $i = 1 , 2,\ldots, N$. Based on this distance matrix, our goal is to accurately locate the agents in a large-scale wireless network with satisfactory computation time.

The above signal model well suits many realistic large-scale wireless networks. For instance, in 5G network, a large number of small base stations are densely deployed in each cell; Internet of Things (IoT) network, advocating interconnection of everything, comprises a huge number of connected smart devices and machines \cite{Yin20fedloc}. For large-scale networks, it is typical that only a small fraction of nodes know their locations precisely. To know all locations, otherwise, one requires either a lot of manpower to do offline calibration or to equip expensive and power-hungry GPS/BeiDou chips.

To locate a large number of agents, we propose a completely new learning paradigm, namely GNN based network localization, which is data-driven and relies on a graph-based machine learning model, to be specified in the next section.

\section{Network Localization with GCN} \label{GCN_reg}

Among different types of GNN models, graph convolutional networks (GCNs) constitute a representative class. In this section, we focus on formulating the network localization problem using GCNs.
An undirected graph associated with a wireless network can be formally defined as ${\mathcal{G}} = ({\mathcal{V}}, \mathbf{A})$, where $\mathcal{V}$ represents the vertex set of the nodes $\{v_1,v_2, \ldots, v_N\}$, and $\mathbf{A}\in\mathbb{R}^{N \times N}$ is a symmetric adjacency matrix where $a_{ij}$ denotes the weight of the edge between $v_i$ and $v_j$. We set $a_{ij} = 1$ if there is no connection between $v_i$ and $v_j$, otherwise $a_{ij} = 0$. The degree matrix $\mathbf{D}\in\mathbb{R}^{N \times N}:= \text{diag}(d_1,d_2, \ldots, d_N)$ is a diagonal matrix with $d_i = \sum_{j = 1}^N a_{ij}$. 

In the original GCN, the edge $a_{ij}$ can be regarded as a similarity coefficient between nodes $i$ and $j$.
In the context of network localization, two nodes are similar means that they are close to each other, i.e., $d(\mathbf{p}_i,\mathbf{p}_j)$ being small. 
Accordingly, we introduce a Euclidean distance threshold, denoted by $T_h$, to determine whether there is an edge between two nodes or not.
As will be explained in Section \ref{analysis}, this threshold is critical to the localization performance.
By thresholding, a refined adjacency matrix $\mathbf{A}_{T_h}\in\mathbb{R}^{N \times N}$ is constructed as follows:
\begin{equation}\label{eq:threshold}
[\mathbf{A}_{T_h}]_{ij}=\left\{\begin{array}{l}
0,\quad \text{if} \quad  x_{ij}>T_h \\
1,\quad \text{otherwise}.
\end{array}\right.
\end{equation}
Consequently, the augmented adjacency matrix \cite{kipf2016semi} is defined as $\tilde{\mathbf{A}}_{T_h}:= \mathbf{A}_{T_h}+\mathbf{I}$, where $\mathbf{I}$ is an identity matrix, and the associated degree matrix of $\tilde{\mathbf{A}}_{T_h}$ is denoted by $\tilde{\mathbf{D}}_{T_h}\in\mathbb{R}^{N \times N}$.

Similarly, we construct a sparse distance matrix
$\hat{\mathbf{X}}=\mathbf{A}_{T_h} \odot \mathbf{X}$, where $\odot$ denotes the Hadamard product. Consequently, $\hat{\mathbf{X}}$ contains only distance measurements that are smaller than or equal to $T_h$.

In general, each layer of GCN carries out three actions: feature propagation, linear transformation and an element-wise nonlinear activation \cite{wu2019simplifying}. 
The main difference between GCN and the standard multi-layer perceptron (MLP) \cite{svozil1997introduction} lies in the feature propagation, which will be clarified in the following.

In the $k$-th graph convolution layer, we assume $D_k$ is the number of neurons in the $k$-th layer, then the input and output node representations are denoted by the matrices $\mathbf{H}^{(k-1)}$ and $\mathbf{H}^{(k)}\in\mathbb{R}^{N \times D_k} $, respectively. The initial node representations is $\mathbf{H}^{(0)} = \hat{\mathbf{X}}$. A $K$-layer GCN differs from a $K$-layer MLP in that the hidden representation of each node in GCN is averaged with its neighbors. 
More precisely, in GCN, the update process for all layers is obtained by performing the following matrix multiplication:
\begin{equation} \label{eq:update_matrix}
\bar{\mathbf{H}}^{(k)}\in\mathbb{R}^{N \times D_{k-1}}  \leftarrow \hat{\mathbf{A}}_{T_h} \mathbf{H}^{(k-1)},
\end{equation}
where $\hat{\mathbf{A}}_{T_h}\in\mathbb{R}^{N \times N} :=\tilde{\mathbf{D}}_{T_h}^{-\frac{1}{2}} \tilde{\mathbf{A}}_{T_h} \tilde{\mathbf{D}}_{T_h}^{-\frac{1}{2}}$ is the augmented normalized adjacency matrix \cite{kipf2016semi} and $\bar{\mathbf{H}}^{(k)}$ is the hidden representation matrix in the $k$-th graph convolution layer. Intuitively, this step smoothes the hidden representations locally along the edges of the graph and ultimately encourages similar predictions among locally connected nodes.

\begin{figure}[t] 
\centering
\includegraphics[width=1\linewidth]{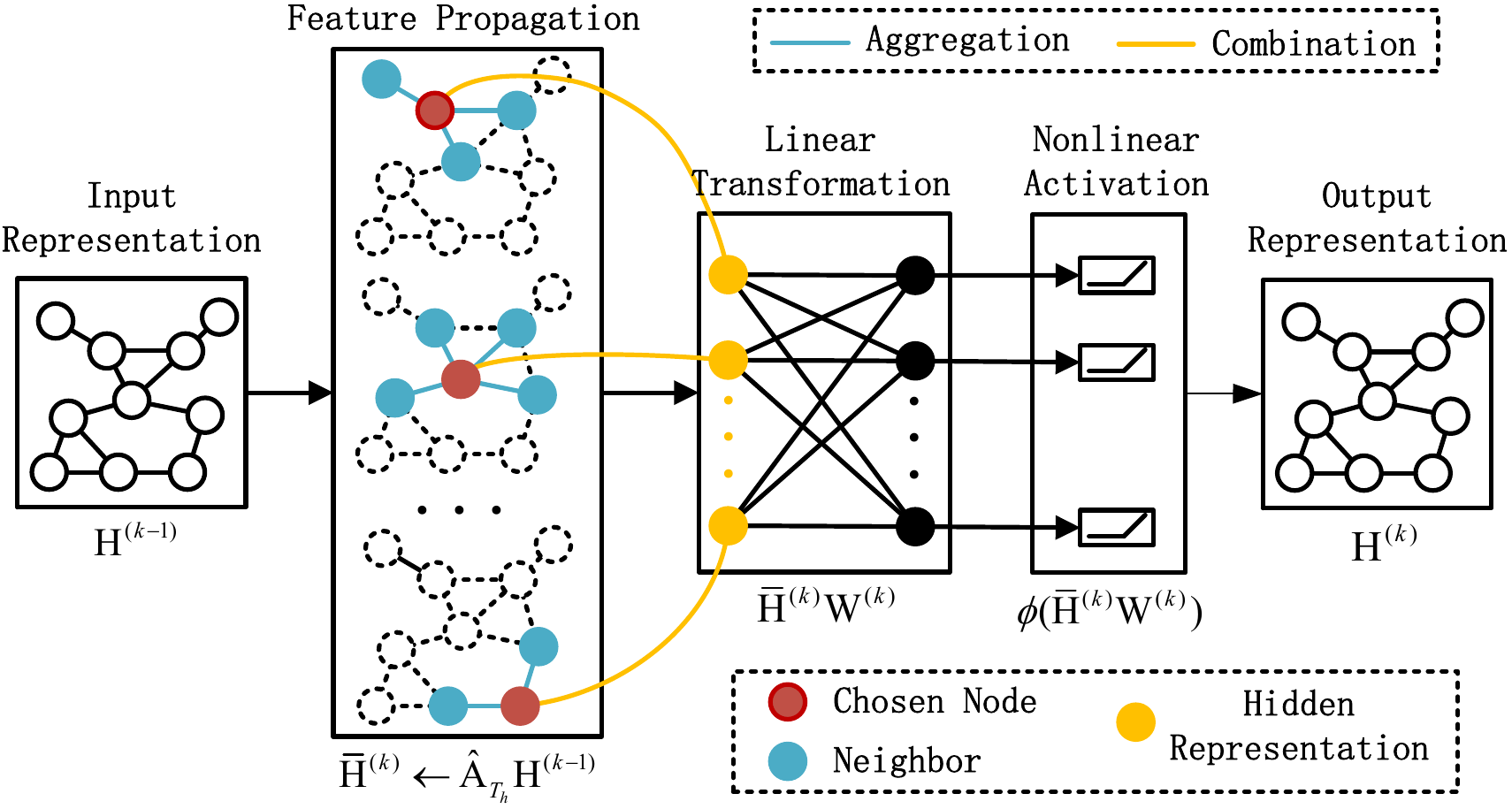}
\setlength{\abovecaptionskip}{-0.3cm}
\caption{Diagram of GCN updating rule for one hidden layer.}
\label{fig:gcn_pictur}
\vspace{-0.3cm}
\end{figure}

After feature propagation, the remaining two steps of GCN, i.e., linear transformation and nonlinear activation, are identical to those of the standard MLP.
The $k$-th layer contains a layer-specific trainable weight matrix $\mathbf{W}^{(k)}\in\mathbb{R}^{D_{k-1} \times D_k} $ and a nonlinear activation function $\phi(\cdot)$, such as $\text{ReLU}(\cdot)= \max(0,\cdot)$ \cite{ramachandran2017searching}. The representation updating rule of the $k$-th layer, presented in Fig.~\ref{fig:gcn_pictur}, is given by
\vspace{-0.3cm}
\begin{align} \label{eq:gcn_propagation}
    \mathbf{H}^{(k)} & \leftarrow  \phi \left( \bar{\mathbf{H}}^{(k)} \mathbf{W}^{(k)}\right). 
\vspace{-0.3cm}
\end{align}
It is noteworthy that the activation function $\phi(\cdot)$ is applied to every element in the matrix. 

Taking a 2-layer GCN as an example, the estimated positions, $\hat{\mathbf{R}}=[\hat{\mathbf{p}}_1,\hat{\mathbf{p}}_2,\dots,\hat{\mathbf{p}}_N]^\top\in\mathbb{R}^{N\times 2}$, are given by
\begin{equation}
\hat{\mathbf{R}} 
=  \hat{\mathbf{A}}_{T_h} \,\, \phi\!\left(\hat{\mathbf{A}}_{T_h} (\mathbf{A}_{T_h} \odot \mathbf{X}) \mathbf{W}^{(1)} \right) \mathbf{W}^{(2)} \, .
\label{eq:two-layer-gcn-for-regression}
\end{equation}
The weight matrices $\mathbf{W}^{(1)}$ and $\mathbf{W}^{(2)}$ can be optimized by minimizing the mean-squared-error (MSE), $\mathcal{L}(\mathbf{W}^{(1)},\mathbf{W}^{(2)}):=\|\mathbf{R}_l-\hat{\mathbf{R}}_l \|^2_F$, where $\mathbf{R}_l = [\mathbf{p}_1,\mathbf{p}_2,\dots,\mathbf{p}_{N_l}]^\top$ and $\hat{\mathbf{R}}_l =  [\hat{\mathbf{p}}_1,\hat{\mathbf{p}}_2,\dots, \hat{\mathbf{p}}_{N_l}]^\top$ are the true anchor positions and their estimates, respectively, and $\|\cdot\|_F$ is the Frobenius norm of a matrix. This optimization problem is often solved via a gradient descent type method, such as stochastic gradient descent \cite{bottou2010large} or Adam \cite{kingma2014adam}.

\section{Numerical Results}\label{experiment}

\begin{table*}
	\centering
	\setlength{\abovecaptionskip}{-0cm}
	\setlength{\belowcaptionskip}{-0.2cm}
	\caption{The averaged loss (RMSE) of all methods under different noise conditions for $N_l$=50.}
	\label{tab:1}       
	\begin{tabular}{cccccc}
		\hline\noalign{\smallskip}
		Methods$\backslash$ Noise $(\sigma^2,p_B)$ & $(0.04,0\%)$ & $(0.1,10\%)$ & $(0.25,10\%)$ & $(0.25,30\%)$ & $(0.5,50\%)$  \\
		\noalign{\smallskip}\hline\noalign{\smallskip}
		GCN  & \textbf{0.1038} & \textbf{0.1128} & \textbf{0.1006} & \textbf{0.1302} & \textbf{0.1755}\\
		$\text{GCN}_{1000}$ & 0.0874 & 0.0856 & 0.0998 & 0.0981 & 0.1404\\
		MLP  & 0.1865 & 0.1769 & 0.2305 & 0.2623 & 0.3358\\
		NTK\cite{arora2019harnessing}  & 0.4307 &  0.5155 & 0.5270 & 0.6154 & 0.9578\\
		SDP\cite{jin2020exploiting}  & 0.1171 & 0.2599 & 0.4891 & 0.4641 & 0.9294\\
		ECM\cite{Yin_ECM}  & 0.1610 & 0.1857 & 0.3298 & 0.3824 & 0.8011\\
		LS\cite{Wymeersch2009}  & 0.2270 & 0.2675 & 0.3884 & 0.4187 & 0.7992\\
		\noalign{\smallskip}\hline
	\end{tabular}
	\vspace{-0.5cm}
\end{table*}

In this section, the performance of the proposed GCN based method is evaluated in terms of localization accuracy, robustness against NLOS noise and computational time. For comparison purposes, we choose various well performed competitors, including an MLP based method, a neural tangent kernel (NTK) regression based method, the sparsity-inducing semi-definite programming (SDP) method \cite{jin2020exploiting}, the expectation-conditional maximization (ECM) method \cite{Yin_ECM}, and the centralized least-square (LS) method \cite{Wymeersch2009}. Note that we choose MLP to demonstrate the performance improvement caused by adding the normalized adjacent matrix $\hat{\mathbf{A}}_{T_h}$ in each layer.
Additionally, we use NTK based regression to mimic ultra-wide MLP with random initialization based on the theorem that a sufficiently wide and randomly initialized MLP trained by gradient descent is equivalent to a kernel regression predictor with the NTK \cite{arora2019exact}. 

Implementation details of these methods are as follows. We use a 2-layer GCN with 2000 neurons in each hidden layer. 
We train GCN and MLP models for a maximum number of 200 epochs (full batch size) using Adam with a learning rate of 0.01. We initialize the weights using the routine described in \cite{glorot2010understanding} and normalize the input feature vectors along rows. Dropout rate \cite{srivastava2014dropout} is set to 0.5 for all layers. 
The settings of NTK remain the same as described in \cite{arora2019harnessing}. 
The regularization parameter in SDP is set to $\lambda = 0.05$. For both the ECM and LS methods, the initial positions are randomly generated in the square area.
All simulations are performed on a server computer equipped with 48 Inter Xeon E5-2650 2.2GHz CPUs and 8 NVIDIA TITAN Xp 12GB GPUs. In all experiments, we set the threshold $T_h=1.2$ for GCN, MLP and NTK, and set $T_h=0.6$ for other methods, which leads to similar averaged localization accuracy but requires much less computational time than using $T_h=1.2$. Fairness in comparison is carefully maintained.

Details of the simulated localization scenarios are given below. We consider a 5m$\times$5m unit square area. Each network consists of $500$ randomly generated nodes in total. The number of anchors, $N_l$, varies from $20$ to $160$ for investigating its impact, and the rest are agents. 
The measurement error $n_{ij}$ is generated according to $n_{ij} = n^L_{ij} + b_{ij}n^N_{ij}$. 
Here, the LOS noise $n^L_{ij}$ is generated from a zero-mean Gaussian distribution, i.e., $n^L_{ij}\sim\mathcal{N}(0, \sigma^2)$, while the positive NLOS bias $n_{ij}^N$ is generated from the uniform distribution, $n_{ij}^N\sim\mathcal{U}[0, 10]$ and $b_{ij}$ generated from the Bernoulli distribution $\mathcal{B}(p_B)$ with $p_B$ being the NLOS occurrence probability.

\begin{figure}[t] 
\centering
\includegraphics[width=1\linewidth]{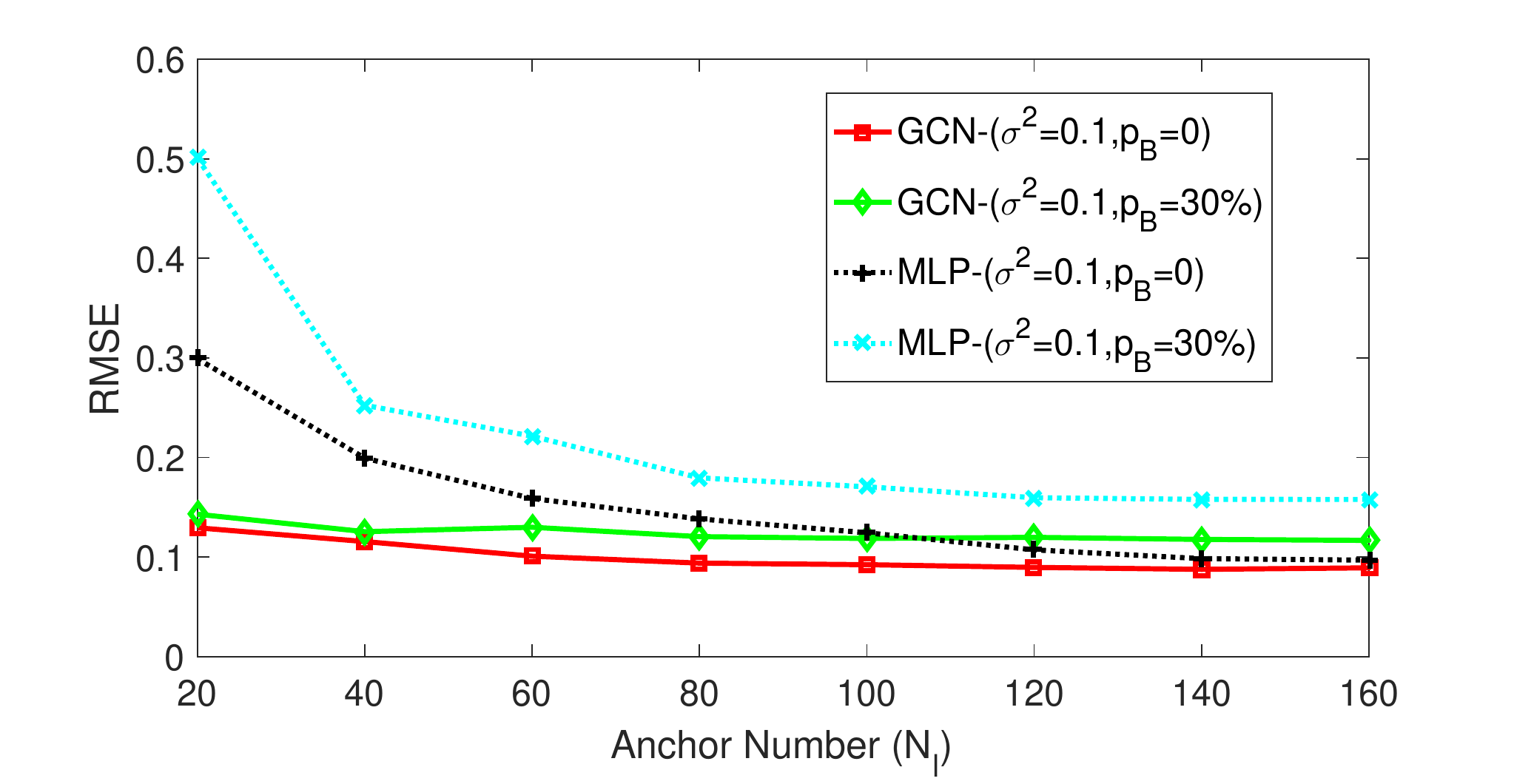}
\setlength{\abovecaptionskip}{-0.3cm}
\caption{The averaged loss (RMSE) versus the number of anchors under different noise conditions.}
\label{fig:anchor}
\vspace{-0.3cm}
\end{figure}

First, we assess the localization accuracy of all methods under different noise conditions. Here, the localization accuracy is measured in terms of the averaged test root-mean-squared-error (RMSE), 
$\mathcal{L}_R:=\|\mathbf{R}_u-\hat{\mathbf{R}}_u \|_F$, where $\mathbf{R}_u = [\mathbf{p}_{N_l+1},\mathbf{p}_{N_l+2},\dots,\mathbf{p}_N]^\top$ and $\hat{\mathbf{R}}_u =  [\hat{\mathbf{p}}_{N_l+1},\hat{\mathbf{p}}_{N_l+2},\dots, \hat{\mathbf{p}}_{N}]^\top$. 
The results are summarized in Table~\ref{tab:1}. It is shown that among all considered methods, GCN provides the highest localization accuracy in almost all cases. In particular, when the NLOS probability, $p_B$, is high, GCN outperforms all competitors by far. Moreover, we test the localization performance of GCN for large networks with $N=1000$, denoted by $\text{GCN}_{1000}$ in Table~\ref{tab:1}. The results show that GCN performs even better with slightly increased computational time, as shown in Table~\ref{tab:2}. If we further increase $N$, the GCN based method can maintain its performance, but the other methods (SDP, ECM and LS) will all degrade severely in terms of localization accuracy and computational time.

Next, we focus on two data-driven methods, GCN and MLP, which perform relatively well in Table~\ref{tab:1}.
The localization accuracy is investigated by varying $N_l$ from $20$ to $160$ with a stepsize of $20$ under two different noise conditions. 
The results are depicted in Fig.~\ref{fig:anchor}. There are two main observations. First, GCN attains the lowest RMSE consistently for all $N_l$. Compared with MLP, the improvement in localization accuracy is particularly remarkable for GCN with small $N_l$. This result indicates that GCN can better exploit the distance information than MLP. When $N_l$ increases, both GCN and MLP tend to approach a performance lower bound. Second, GCN performs similarly under both noise conditions, while MLP shows a clear performance degradation when $p_B$ increases. This observation indicates that GCN is very robust against NLOS. Lastly, we want to mentioned that a fine-tuned MLP is often superior to NTK which corresponds to random initialized MLP, which performs surprisingly close to other benchmark methods as shown in Table~\ref{tab:1}.

\begin{figure}[t] 
\centering
\includegraphics[width=1\linewidth]{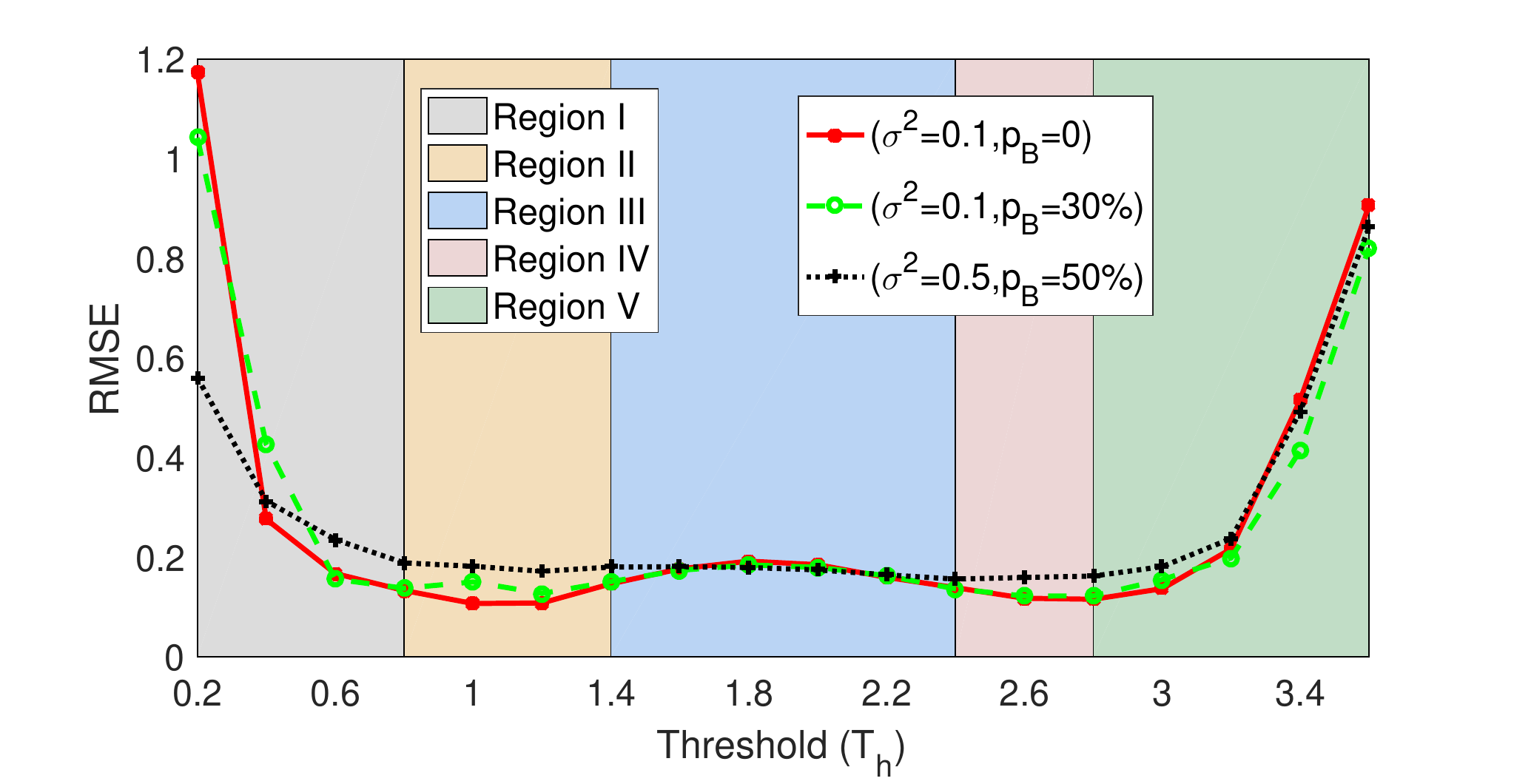}
\setlength{\abovecaptionskip}{-0.3cm}
\caption{The averaged loss (RMSE) versus threshold under different noise conditions in GCN model. $N_l = 50$.}
\label{fig:threshold}
\vspace{-0.3cm}
\end{figure}

In the third simulation, we focus on investigating the influence of the threshold, $T_h$, on the localization performance. Figure~\ref{fig:threshold} depicts the RMSE of GCN versus the threshold, $T_h$, in three different noise scenarios. 
It is interesting to see that the RMSE curves show similar trend as the threshold changes. We characterize the localization performance obtained by using different thresholds. In \textbf{Region I} ($T_h\in[0.2,0.8]$), the RMSE is very large at the beginning and drops rapidly as $T_h$ increases. The reason for such bad performance at the beginning is that when $T_h$ is too small there will be no sufficient edges in the graph, incurring isolated nodes. In \textbf{Regions II} $\sim$ \textbf{IV} ($T_h\in(0.8,2.8]$), GCN shows stable performance. A closer inspection shows that the RMSE is relatively lower in \textbf{Region II}, rises slightly in \textbf{Region III} and decreases in \textbf{Region IV} to the same lowest level as in \textbf{Region II}. This observation can be explained as follows. When $T_h\in(0.8,1.4]$, the good performance of GCN is due to the NLOS noise truncation effect of $T_h$, which will be explained in Section~\ref{analysis_threshold}. For $T_h\in(2.4,2.8]$, the adverse effect of large NLOS noise is compensated by the increased number of neighbors for each node. Lastly, the rapid increase of RMSE in \textbf{Region V} can be explained by the effect of extremely large NLOS noise and over-smoothing, which will be explained in Section~\ref{analysis_threshold} as well.

Another important requirement for real-world applications is fast response. Table~\ref{tab:2} shows the practical computational time of different methods. It is shown that GCN, MLP and NTK are more computationally efficient than the other methods.
Besides, the computational time of $\text{GCN}_{1000}$  only slightly increases when we double the number of nodes in the network. Notably, GCN can process very large network, for instance $N=10000$, in an affordable time, while the LS, ECM and SDP are all disabled in this case. 
All above results indicate that the proposed GCN based method is a prospective solution to large-scale network localization.

\section{Performance Reasoning}
\label{analysis}
In this section, we aim to dig out the reasons behind the remarkable performance of the newly proposed GCN based method, corroborated by the results shown in Section \ref{experiment}. 
We pinpoint two major factors: one is the threshold $T_h$ and the other is the augmented normalized adjacency matrix $\hat{\mathbf{A}}_{T_h}$. In the following, we analyze the two factors separately, although $T_h$ determines $\hat{\mathbf{A}}_{T_h}$.

\subsection{Effects of Thresholding} 
\label{analysis_threshold}
Thresholding plays two major roles: 1) truncating large noise and 2) avoiding over-smoothing.

\vspace{4pt}
\noindent
\textbf{Noise truncation.} 
The operation, $\hat{\mathbf{X}}=\mathbf{A}_{T_h} \odot \mathbf{X}$, in Section~\ref{GCN_reg} implies that for each $\hat{x}_{ij}\neq 0$, $
d(\mathbf{p}_i,\mathbf{p}_j)+n_{ij}\leq T_h$ holds.
Equivalently speaking, for each non-zero element in $\hat{\mathbf{X}}$, we have $n_{ij}\leq T_h-d(\mathbf{p}_i,\mathbf{p}_j)$, indicating that only noise in a small limited range will be included in $\hat{\mathbf{X}}$. Specifically, due to the fact that $n_{ij}$ with large value is usually caused by positive NLOS bias $n_{ij}^{N}$, each element $x_{ij}$, associated either with a large true distance or with a large noise, is neglected when the threshold $T_h$ is set small. In other words, we retain the measurement if the two nodes are adjacent and only affected by a small or moderate measurement noise.

\begin{table}
\setlength{\abovecaptionskip}{-0cm}
\setlength{\belowcaptionskip}{-0.2cm}
\setlength\tabcolsep{4pt}
\footnotesize
\centering
\caption{A comparison of different methods in terms of computational time (in second) at $(\sigma^2=0.1, p_B=30\%)$ and $N_l = 50$.}
\label{tab:2}       
\begin{tabular}{cccccccc}
\hline\noalign{\smallskip}
 GCN & $\text{GCN}_{1000}$ & $\text{GCN}_{10000}$ & MLP & NTK & LS & ECM & SDP  \\
\noalign{\smallskip}\hline\noalign{\smallskip}
 3.24 & 5.82 & 707.38 & \textbf{2.05} & 2.33 & 32.47 & 82.85 & 1587  \\
\noalign{\smallskip}\hline
\end{tabular}
\vspace{-0.3cm}
\end{table}

\vspace{4pt}
\noindent
\textbf{Avoiding over-smoothing.} When the threshold is large enough, the corresponding graph will become fully connected and the adjacency matrix will be a matrix of all ones. In this case, according to Eq.~\eqref{eq:update_matrix}, all entries of the hidden representation matrix $\bar{\mathbf{H}}^{(k)}$ are identical, meaning that the obtained hidden representation completely loses its distance information. Consequently, the predicted positions of all nodes will tend to converge to the same point. This phenomenon is known as over-smoothing. As an illustration, \textbf{Region V} in Fig.~\ref{fig:threshold} confirms that GCN will suffer from over-smoothing when the threshold is set too large. Thus, we need to choose a proper threshold to prevent such an adversarial behavior.

\subsection{Effects of $\hat{\mathbf{A}}_{T_h}$}
To understand the superior localization performance of GCN compared with MLP, we analyze the effects of the augmented normalized adjacency matrix, $\hat{\mathbf{A}}_{T_h}$, from both spatial and spectral perspectives.

\vspace{4pt}
\noindent
\textbf{Aggregation and combination.} To understand the spatial effect of $\hat{\mathbf{A}}_{T_h}$, we decompose $\hat{\mathbf{A}}_{T_h}$, cf. Eq.~\eqref{eq:update_matrix}, into two parts:
\vspace{-0.2cm}
\begin{equation}
    \bar{\mathbf{h}}_i^{(k)} = \underbrace{\frac{1}{d_i + 1} \mathbf{h}_i^{(k-1)}}_{\text{Own information}}+\underbrace{\sum_{j=1}^{N} \frac{a_{ij}}{\sqrt{(d_i + 1) (d_j + 1)}}\mathbf{h}_j^{(k-1)}}_{\text{Aggregated information}},
\label{eq:update}
\vspace{-0.3cm}
\end{equation}
where $\bar{\mathbf{h}}_{i}^{(k)}$ and $\mathbf{h}_i^{(k)}$ are the $i$-th row vectors of hidden representation matrix, $\bar{\mathbf{H}}^{(k)}$, and the input representation matrix, $\mathbf{H}^{(k)}$, in the $k$-th layer, respectively. 
Specifically, Eq.~\eqref{eq:update} contains two operations: aggregation and combination. Aggregation corresponds to the second term of Eq.~\eqref{eq:update}, in which the neighboring features are captured by following the given graph structure. Then, the target node's own information is combined with the aggregated information. 

Comparing with the training procedure of MLP, which solely uses the features of the labeled nodes (referred to as anchors here), GCN is a semi-supervised method in which the hidden representation of each labeled node is averaged for a carefully tailored local neighborhood including itself. Equivalently speaking, GCN trains a model by exploiting features of both labeled and unlabeled nodes, leading to superior localization performance.

\begin{figure}[tb] 
\centering
\includegraphics[width=1\linewidth]{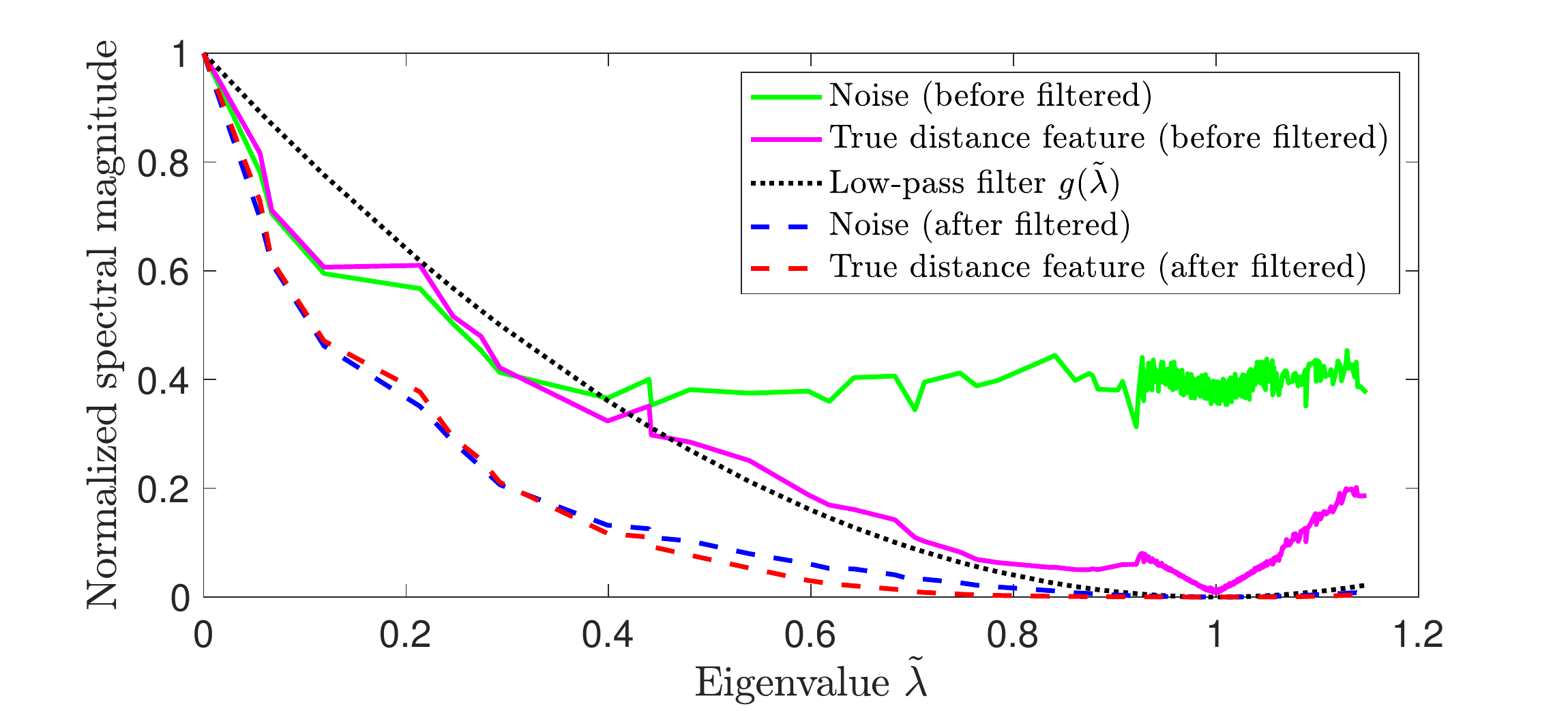}
\setlength{\abovecaptionskip}{-0.3cm}
\caption{Spectral components of different signals in dataset $(\sigma^2=0.1,p_B=0)$.}
\label{fig:frequency_plot}
\vspace{-0.3cm}
\end{figure}

\vspace{4pt}
\noindent
\textbf{Low-pass filtering.} From the spectral perspective, the eigenvalues of the augmented normalized Laplacian $\mathbf{L}_{T_h} = \mathbf{I}-\hat{\mathbf{A}}_{T_h}$, denoted by $\tilde{\lambda}$, can be regarded as the "frequency" components \cite{sandryhaila2014discrete, gama2020graphs}. Multiplying $K$ augmented normalized adjacency matrices $\hat{\mathbf{A}}^K_{T_h}$ in graph convolution layers is equivalent to passing a spectral ``low-pass'' filter $g(\tilde{\lambda}_i) := (1 - \tilde{\lambda}_i)^K$, where $\tilde{\lambda}_i,i=1,2,\ldots$ \cite{wu2019simplifying}.
Figure~\ref{fig:frequency_plot} depicts the ``frequency'' components of the LOS noise and the true distance matrix before and after the filtering process. 
It can be seen that almost all information of the true distance matrix (before the filtering)  is concentrated in the ``low frequency'' band, while both ``low frequency'' and ``high frequency'' components are present in the LOS noise before the filtering. Thus, $\hat{\mathbf{A}}_{T_h}$, acting as a ``low-pass'' filter, can partially remove the ``high frequency'' component of the LOS noise. This explains the improved localization performance of GCNs from the spectral perspective.

\section{Conclusion}\label{conclusion}
In this paper, we have proposed a GCN based data-driven method
for robust large-scale network localization in mixed LOS/NLOS environments. 
Numerical results have shown that the proposed method
is able to achieve substantial improvements in terms of localization accuracy, robustness and computational time, in comparison with both MLP and various state-of-the-art benchmarks. Moreover, our detailed analyses found that thresholding the neighboring features is crucial to attaining superb localization performance. The proposed data-driven paradigm is believed to drive more efficient and robust methods for network localization and related ones in the future.

\bibliographystyle{IEEEbib}
\bibliography{strings,refs}

\end{document}